\documentclass[conference]{IEEEtran}
\usepackage{etoolbox}
\makeatletter
\patchcmd{\maketitle}{\@copyrightspace}{}{}{}
\usepackage{comment}
\usepackage{float}
\usepackage{wrapfig}

\usepackage{psfrag}  
\usepackage{subfigure}
\usepackage{epsfig}
\usepackage{url}
\usepackage{amsmath, amssymb}
\usepackage{array}

\usepackage[utf8x]{inputenc}
\usepackage{lipsum}
\usepackage{algpseudocode}
\usepackage{siunitx} 
\usepackage{fancyhdr,graphicx,amsmath,amssymb}
\usepackage{xspace}
\usepackage[T1]{fontenc}
\include{pythonlisting}
\usepackage{color,soul}
\usepackage{fixltx2e}
\usepackage{hyperref}
\usepackage{cleveref}
\usepackage[T1]{fontenc}
\usepackage{ifpdf}
\usepackage{algorithm} 
\usepackage{algpseudocode}  
\usepackage{graphicx}
\usepackage{multirow}
\usepackage{booktabs}

\begin{document}

\title{\LARGE KV-CAR: \underline{KV} Cache \underline{C}ompression using \underline{A}utoencoders and KV \underline{R}euse in Large Language Models}



%

\author{\IEEEauthorblockN{Sourjya Roy, Shrihari Sridharan, Surya Selvam, Anand Raghunathan}
\IEEEauthorblockA{Elmore Family School of Electrical and Computer Engineering, Purdue University, West Lafayette, IN}}


\maketitle

\begin{abstract}
As Large Language Models (LLMs) scale in size and context length, the memory 
requirements of the key--value (KV) cache have emerged as a major bottleneck 
during autoregressive decoding. The KV cache grows with sequence 
length and embedding dimension, often exceeding the memory footprint of the model 
itself and limiting achievable batch sizes and context windows.

To address this 
challenge, we present KV-CAR, a unified and architecture-agnostic framework that 
significantly reduces KV-cache storage while maintaining model fidelity.
KV-CAR combines two complementary techniques. First, a lightweight autoencoder learns compact representations of key and value tensors along the embedding dimension, compressing them before they are stored in the KV cache and restoring them upon retrieval. Second, a similarity-driven reuse mechanism identifies 
opportunities to reuse KV tensors of specific attention heads across adjacent layers. Together, these methods reduce the dimensional and structural redundancy in KV tensors without requiring changes to 
the transformer architecture.
Evaluations on GPT-2 and TinyLLaMA models across Wikitext, C4, PIQA, and Winogrande datasets
demonstrate that KV-CAR achieves up to 47.85\% KV-cache memory 
reduction with minimal impact on perplexity and zero-shot accuracy. System-level measurements on an NVIDIA A40 GPU show that the 
reduced KV footprint directly translates into longer sequence lengths and larger batch sizes during inference. These results highlight the effectiveness of KV-CAR in enabling memory-efficient LLM 
inference.
\end{abstract}

\IEEEoverridecommandlockouts


%
\IEEEpeerreviewmaketitle

\section{Introduction}
Large Language Models (LLMs) have achieved remarkable performance across a wide range of natural language and multimodal tasks due to their ability to capture long-range dependencies and generate contextually rich outputs. This capability, however, is accompanied by substantial computational and memory requirements. Modern LLMs routinely contain billions of parameters, and empirical scaling laws indicate that model quality continues to improve as models grow larger—placing increasing pressure at inference time on compute and memory subsystems.

Most state-of-the-art LLMs use a decoder-only transformer and perform inference in two phases: \emph{prefill} and \emph{decode}. In the prefill phase, the model processes the full input prompt and produces the key and value vectors for all tokens across all layers. These vectors are stored in a structure known as the \emph{KV cache}. In the decode phase, tokens are generated autoregressively: each new token attends to all previously cached representations, and its own key and value vectors are appended to the cache.

Although these key and value tensors could, in principle, be recomputed at every decoding step, doing so would require repeating the full sequence of KV projections across multiple layers for every previously generated token. As sequence length and batch size increase, the amount of repeated computation becomes prohibitively large. To avoid this cost, current systems cache and retrieve the KV tensors during decoding. While this significantly reduces compute, it introduces a major memory burden: the KV cache grows in proportion to the product of the sequence length, batch size and embedding dimension of each attention layer. As a result, the KV cache often becomes the dominant contributor to inference-time memory consumption on GPU systems, and it directly limits the \emph{maximum supported batch size} and the \emph{maximum achievable context length}. In practice, the cache can reach tens of gigabytes even for moderate sequence lengths and batch sizes, causing out-of-memory failures despite available compute headroom.

As LLMs scale toward longer context lengths and high-throughput batched decoding, managing KV-cache memory has become a central challenge for efficient inference. Several strategies have been adopted in attempts to mitigate this bottleneck. KV quantization reduces precision of entries in the KV cache, while token-pruning removes less influential tokens. The model can be redesigned to share attention projections across heads or layers. Offloading solutions move portions of the KV cache to host memory at the cost of additional latency during retrieval. While these approaches have proven somewhat effective, the KV cache remains an ongoing bottleneck with the drive towards very long context lengths. Critically, prior approaches do not reduce the embedding dimensionality of key and value vectors—a key factor in KV-cache growth. This observation motivates our work, which represents an orthogonal axis of compression that targets the internal structure of KV representations themselves.

In this paper, we propose KV-CAR, a framework that directly reduces KV-cache storage without requiring model redesign or extensive training. Our contributions are threefold:

\begin{itemize}
    \item \textbf{Layer-wise autoencoder compression.} We introduce lightweight per-layer autoencoders that compress full-dimensional key and value vectors from dimension $D$ to a compact latent dimension $d \ll D$ before storage in the KV cache. A decoder reconstructs the vectors upon retrieval before they are used in attention computation. This learned nonlinear mapping adapts to the statistical structure of each layer, offering substantially greater memory savings than fixed linear projections while maintaining accuracy.

    \item \textbf{Similarity-guided attention-head reuse.} We identify redundant attention heads across adjacent layers using an L1-norm similarity metric and reuse their key and value tensors when similarity is high. This reduces the number of stored KV tensors without degrading performance and complements embedding compression by eliminating redundancy at the head level.

    \item \textbf{Lightweight autoencoder training methodology.} We design a stable training pipeline that first trains layer-specific autoencoders independently while keeping the base model frozen, followed by a joint optimization stage for selected compressed layers using a hybrid reconstruction--cross-entropy objective. This staged process ensures convergence and preserves downstream task accuracy.
\end{itemize}

Evaluations on the GPT-2 and TinyLLaMA models across the Wikitext, C4, PIQA, and Winogrande datasets show that KV-CAR achieves up to $47.85\%$ KV-cache memory reduction with minimal impact on perplexity and zero-shot accuracy. System-level analysis on an NVIDIA A40 GPU confirms that shrinking the KV cache directly enables \emph{longer context lengths} and \emph{larger batch sizes} before memory exhaustion. These results suggest that KV-CAR can significantly extend the scalability of LLM inference on memory-constrained GPU systems.

\section{Background} \label{sec:background}
In this section, we present an overview of multi-head attention followed by a description of the KV cache and key parameters that impact its memory requirements.

\subsection{Multi-head Attention}

Initially developed for natural language processing (NLP), attention mechanisms have transformed deep learning for various modalities by enabling models to focus on key parts of input sequences dynamically. Among these, multi-head attention (MHA) is pivotal, especially in transformer architectures, as it allows models to attend to multiple aspects of an input simultaneously. MHA operates through several independent "heads," each focusing on different subspaces of the input, thus capturing a wider range of contextual relationships.

Mathematically, MHA consists of multiple scaled dot-product attention heads. Each head computes the attention scores by computing dot producs between queries (\textbf{Q}) and keys (\textbf{K}). These are scaled based on the key dimension, passed through a softmax operation, and multiplied by the values (\textbf{V}). These outputs are concatenated and linearly transformed, yielding a final representation that integrates various dependencies across the sequence. The attention formula for a single head is given as: \begin{equation} \text{Attention}(\textbf{Q}, \textbf{K}, \textbf{V}) = \text{softmax}\left(\frac{\textbf{Q}\textbf{K}^T}{\sqrt{d_k}}\right)\textbf{V} \end{equation} and for MHA: \begin{equation} \text{MultiHead}(\textbf{Q}, \textbf{K}, \textbf{V}) = \text{Concat}(\text{head}_1, \ldots, \text{head}_h)\textbf{W}^O \end{equation} where $\textbf{W}^O$ is a learnable weight matrix. MHA enhances the ability of models to capture both long-range dependencies and local interactions, making it a key component of transformers and all modern LLMs.

\subsection{KV-cache}

LLMs present significant computational challenges during inference, primarily due to the high cost of repeated attention computations. To mitigate this, transformer architectures employ the \textit{Key–Value (KV) cache} mechanism, which stores the key (K) and value (V) matrices for each attention layer and reuses them in subsequent decoding steps. This caching strategy eliminates redundant matrix computations for previously processed tokens, thereby improving inference efficiency.

By reusing the stored K and V matrices, the model reduces inference time complexity from $O(n^2)$ to $O(n)$, where $n$ denotes the sequence length, since attention is computed only for newly generated tokens. However, this optimization introduces a significant \textit{memory overhead}, as the KV cache grows in proportion to the product of several parameters.
The total memory required to store the KV cache can be expressed as:
\begin{equation}
\label{eq:kvsize}
\text{KV\_Cache\_Size} = 2 \times P \times N_{\text{layers}} \times d_{\text{model}} \times L_{\text{seq}} \times B
\end{equation}
where the factor of 2 accounts for the need to store both the key and value tensors, $P$ represents the number of bytes per element (e.g., 2 bytes for FP16), $N_{\text{layers}}$ is the number of transformer layers, $d_{\text{model}}$ is the embedding dimension, $L_{\text{seq}}$ is the sequence length, and $B$ is the batch size. This equation highlights the dependency of KV cache size on sequence length, model width, and batch size, illustrating why long-context inference quickly becomes memory-bound.

To provide a concrete example, consider the GPT-2 Medium model, which has $N_{\text{layers}} = 24$ and $d_{\text{model}} = 1024$. Assuming FP16 precision (2 bytes), a sequence length $L_{\text{seq}} = 2048$, and a batch size $B = 8$, the total KV cache size is:
\[
\text{KV\_Cache\_Size} = 2 \times 2~\text{bytes} \times 24 \times 1024 \times 2048 \times 8 \approx 1.61~\text{GB}.
\]
By comparison, the model itself has approximately 345 million parameters (roughly 690~MB in FP16). Thus, during long-context inference, the KV cache alone can occupy around $2.33\times$ the memory of the model parameters.
Efficient management of the KV cache is therefore crucial to balancing inference speed, memory usage, and overall performance as LLMs continue to scale. We discuss prior efforts to address this challenge in the next section and place our work in their context.

\section{Related Work} \label{sec:relatedwork}
Existing methods to address the KV cache bottleneck during LLM inference include KV quantization, token pruning, architectural changes to lower the number of KV heads, and memory offloading techniques. We outline these approaches and discuss how our approach differs from and complements them below.

\textbf{Quantization-Based KV Compression.} Quantization reduces KV-cache memory by lowering the precision of stored tensors without altering the tensor dimensions~\cite{awq2024, dettmers2023gptq, lin2023awq, rotatekv2024}. Prior work has shown that KV tensors can often be represented at low bitwidths with minimal impact on downstream accuracy. While useful, the push towards longer context lengths and batch sizes that are sufficiently large to utilize the computational resources of the inference hardware platform
implies that KV cache remains a bottleneck. We note that quantization operates purely along the precision axis and does not reduce the dimensionality of the KV vectors themselves. Thus, our approach is complementary to and can be combined with quantization. 

\textbf{Dynamic and Attention-Based Pruning.} These works reduces KV-cache memory by pruning tokens based on their estimated importance during decoding. Attention-driven approaches such as Keyformer~\cite{adnan2024keyformer} retain only those tokens that account for the majority of attention mass, discarding the rest to shrink the cache along the sequence dimension. More recent adaptive schemes, including ~\cite{zeng2025lethe}, vary pruning budgets across layers and time by using relevance aware heuristics to remove tokens dynamically. Other methods such as ~\cite{mustafar2025} apply unstructured sparsity directly to KV entries, selecting a compact subset of tokens that contribute meaningfully to downstream predictions. These techniques reduce memory by reducing the sequence length, whereas our method compresses the representation of each token. As a result, pruning-based strategies and our embedding-dimension compression are complementary and can be combined for additional KV-cache savings.

\textbf{Architectural Modifications.} Another class of techniques reduces KV-cache memory by altering the attention architecture itself. Multi-Query Attention (MQA)~\cite{shazeer2019mqa} and Grouped-Query Attention (GQA)~\cite{ainslie2023gqa} share key and value projections across multiple query heads, lowering the number of KV heads that must be stored during decoding. More recent variants~\cite{brandon2024crosslayerkv} extend this idea across layers, allowing deeper layers to reuse KV projections computed earlier in the network and thereby reducing the total number of distinct KV tensors. Other works~\cite{cai2024pyramidkv} restructure the transformer into hierarchical or pyramidal forms that allocate larger KV budgets to early layers while aggressively compressing deeper ones. These architectural approaches reduce memory by decreasing the quantity of KV heads generated in the model. Our method differs in that we compress the dimensionality of each KV vector. As a result, our technique is compatible with these modified architectures and provide additional benefits.

\textbf{Memory Offloading techniques.} These approaches address KV-cache memory by offloading some portions to CPU or host memory, paging segments in and out as needed during decoding. Methods such as KV-cache paging and streaming systems~\cite{zhao2024paging} maintain a working set while keeping the remainder in lower-cost memory, enabling longer contexts without modifying the model. More recent works integrate token importance estimation to guide which KV blocks remain on-device, improving bandwidth utilization during long-sequence generation~\cite{lee2024infinigen}. These approaches trade increased transfer latency for substantial memory savings. Offloading reduces the placement cost of KV tensors rather than their size. Our method is complementary in that the embedding-dimension compression can be applied before offloading, decreasing both memory pressure and the volume of data transferred across devices.

\vspace{0.8em}
\begin{table*}[htb]
\centering
\caption{Comparison of key-value (KV) cache compression and optimization techniques for transformer inference. 
Our work focuses on learned embedding-dimension compression and intra-layer KV reuse, 
distinct from static, quantization, or architecture-dependent approaches.}
\label{tab:kv_compression}

\small
\setlength{\tabcolsep}{6pt} 
\renewcommand{\arraystretch}{1.25}

\begin{tabular}{@{}p{0.16\textwidth} p{0.20\textwidth} p{0.28\textwidth} p{0.28\textwidth}@{}}
\toprule
\textbf{Category} & \textbf{Method} & \textbf{Compression Strategy} & \textbf{Limitations} \\ \midrule

\textbf{Quantization-Based} &
GPTQ, AWQ~\cite{dettmers2024gptq, awq2024} &
Quantize KV tensors to int8/fp8 precision &
Limited gain over fp16; sensitive to rounding error \\

\textbf{Architectural Compression} &
Grouped Query Attention (GQA)~\cite{shazeer2023gqa} &
Share K/V across query heads (2--4$\times$ reduction) &
Requires architectural modification \\

\textbf{Low-Rank / Linear Projection} &
Mistral~\cite{leviathan2024mistral} &
Project KV tensors into a low-rank subspace &
Fixed-rank bottleneck may reduce context fidelity \\

\textbf{Dynamic / Attention-Based} &
KV Pruning~\cite{li2024kvprune} &
Drop low-attention tokens to reduce storage &
Risk of quality degradation on long-context reasoning \\

\textbf{Memory Offloading} &
KV Paging / Offloading~\cite{zhao2024paging} &
Store and prefetch KV tensors between host and device &
No compression; performance is hardware-dependent \\

\textbf{This Work: Learned Compression + Reuse} &
\textbf{Autoencoder-based KV compression with inter-layer reuse} &
\textbf{Learned embedding-dimension reduction plus structural KV reuse guided by L1-distance and CE-based finetuning} &
\textbf{First joint approach combining KV embedding compression with cross-layer cache reuse} \\

\bottomrule
\end{tabular}
\end{table*}

\section{Methodology} \label{sec:Methodology}
\begin{figure*}[t]
  \vspace*{-0pt}
  \centering
  \label{fig:overall}\includegraphics[width=\textwidth]{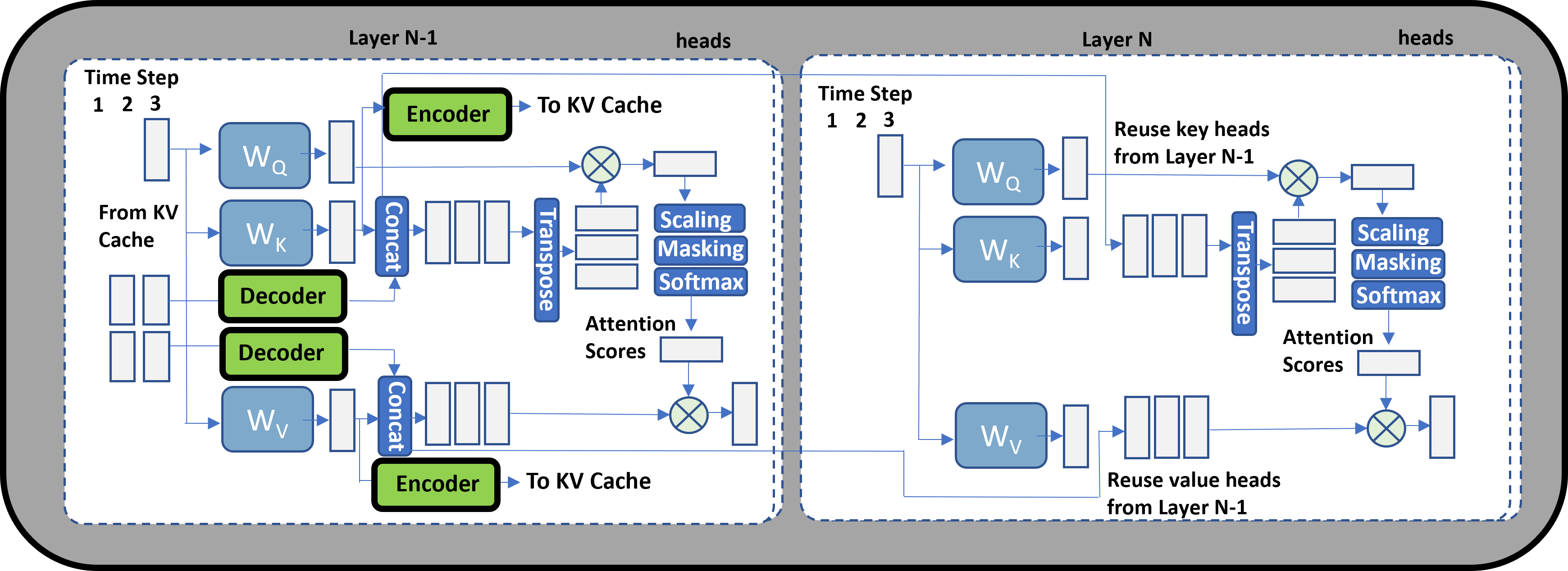}
  \vspace*{-2pt}
  \caption{Block diagram of the KV-CAR framework. For each transformer layer, key and value representations are compressed using a learned autoencoder before insertion into the KV cache. Additionally, structurally redundant attention heads are identified and reused across adjacent layers, jointly minimizing KV-cache storage requirements while preserving inference fidelity.}
  \label{fig:overall}
  \vspace*{-8pt}
\end{figure*}
This section outlines the KV cache compression methodology for two different compression techniques.  Figure \ref{fig:overall} illustrates the multi-head attention mechanism during the decoding phase. Each decoder layer is composed of a multi-head attention module and a feedforward network (FFN) module, with the same decoder structure replicated across multiple layers in the network. Only the multi-head attention module, which is relevant to KV cache quantization, is depicted in Fig \ref{fig:overall}. In the decoding stage, the output of the attention module is passed to the FFN, typically consisting of fully connected layers, activation functions like GELU, and batch normalization. The output of the FFN is then propagated to subsequent layers in the network.
At each time step $t$, the multi-head attention module receives a decoder input of size $D \times 1$, where $D$ represents the embedding dimension. This input is projected through three different weight matrices: $W_Q$, $W_K$, and $W_V$, each of size $D \times D$, to produce corresponding query, key, and value vectors. These vectors are then split into $h$ heads, where each head has a dimension of $\frac{D}{h} \times 1$.The key vectors from the current time step are concatenated with those from the previous $l-1$ time steps (dimension $(l-1) \times D$) and then multiplied with the query vectors, followed by a softmax operation to compute attention scores for each head. The attention outputs are then multiplied by the current and previous value vectors (also of size $(l-1) \times D$). Finally, the attention outputs from all heads are concatenated and passed through a projection matrix of size $D \times D$. This result is normalized using a layer normalization operation, producing an output of size $D \times l$ for the multi-head attention module. The changes in the network Architecture and the training methodology to support the KV cache compression is highlighted in the following subsections. 

\subsection{Network Architecture} \label{sec:networkarch}
A small modification has been introduced to the model architecture, as shown in Fig. \ref{fig:overall}. Specifically, an encoder layer is added after the generation of key and value tokens. This encoder processes the key and value vectors, each originally having dimensions of $1 \times D$, and reduces them to a lower dimension $d$, where $d < D$. The resulting key and value vectors now have dimensions $1 \times d$ and are stored in the key-value (KV) cache located in the high-bandwidth memory (HBM). At each time step $t$, a $1 \times d$ vector is appended to the KV cache for both key and value data structures. 

When retrieving information from the cache, a decoder is applied just before concatenating the previously generated key and value vectors, restoring them to their original dimensions. As a result, at each time step $l$, the $(l-1) \times d$ dimensional key and value matrices are passed through two decoders, expanding them to $(l-1) \times D$. This expanded matrix is then concatenated with the current $1 \times D$ key and value vectors. The autoencoder used in this work consist of an encoder and decoder. The encoder layer consists of two fully connected layers, starting with the $D$-dimensional input. The first fully connected layer may have a different number of neurons than the second layer, which has $d$ neurons. Between these layers, a batch normalization and a Leaky ReLU activation function are applied. The decoder mirrors the encoder, but with the inverse structure: it takes a $d$-dimensional vector as input and outputs $D$ dimensions.

Additionally, a second optimization has been incorporated into the data flow, which can be used independently or in conjunction with the autoencoder compression. In this optimization, certain key and value heads in layer $N$ reuse key and value heads from the previous layer N-1. The heads to be replaced are identified by computing the L1 norm between consecutive layers, which captures the absolute differences between their key and value matrices. A threshold is empirically determined to identify the heads that can be safely replaced without degrading performance. This approach is depicted in the second layer in Fig. \ref{fig:overall}, and represents an inter-layer optimization. Ideally, replacing all the key and value heads between consecutive layers could halve the KV cache requirements. However, in practice, this leads to a drop in application-level accuracy, as discussed in the Results section.
\subsection{Training Methodology}
\begin{algorithm}[htb]
    \caption{Finetuning Procedure for using autoencoders}
    \label{algo:autoencoders}
    \begin{algorithmic}[1]
        \State Initialize model parameters $\theta$ taking a pretrained model
        \State Set learning rate $\alpha$, batch size $B$, number of epochs $E$ and number of layers L
        \State Freeze the gradients for all the layers except the autoencdoer being trained
        \For{$layer= 1$ to $L$}
            \For{$epoch = 1$ to $E$}
                \State Shuffle training data
                \For{each mini-batch $b = 1$ to $N/B$}
                    \State Sample mini-batch of data $\{x_i, y_i\}_{i=1}^{B}$
                    \State Compute predictions $\hat{y}_i = f_\theta(x_i)$
                    \State Compute extra loss as L1 norm of real 
                    \State and predicted key and value for layer 
                    \State Scale the extra loss
                    \State Compute loss $L = \frac{1}{B} \sum_{i=1}^{B} \ell(\hat{y}_i, y_i)$+extraloss
                    \State Compute gradients $\nabla_\theta L$
                    \State Update the autoencoder parameters 
                    \State $\theta = \theta - \alpha \nabla_\theta L$
                   
                \EndFor
            \EndFor
        \EndFor
        \State Return trained parameters $\theta$
        \State The individual autoencoder for all layers have been stored separately
        \State Include autoencoder in the desired layers and initialize with the stored weights
        \State Initialize model parameters $\theta$
        \State Set learning rate $\alpha$, batch size $B$, number of epochs $E$ 
        \State Freeze the gradients for all the layers except all the encoders
        \State Repeat the finetuning process by adding sum of scaled L1 norm loss for all encoders to the final cross entropy loss
        \State Return trained parameters

    \end{algorithmic}
\end{algorithm}
\begin{algorithm}[htb]
    \caption{Finetuning Procedure for inter layer key and value vector reuse}
    \label{algo:kvresue}
    \begin{algorithmic}[1]
        \State Collect the key and value heads for different batches over 1 epoch for different layers
        \State Calculate inter layer L1 norm between heads of adjacent layers
        \State Based on L1 norm, the key and value heads in the particular layers are reused by the key and value in the previous layer
        \State Initialize model parameters $\theta$
        \State Set learning rate $\alpha$, batch size $B$, number of epochs $E$ and number of layers L
        
        \State Initialize model parameters $\theta$
        \State Set learning rate $\alpha$, batch size $B$, number of epochs $E$

        \For{$epoch = 1$ to $E$}
            \State Shuffle training data
            \For{each mini-batch $b = 1$ to $N/B$}
                \State Sample mini-batch of data $\{x_i, y_i\}_{i=1}^{B}$
                \State Compute predictions $\hat{y}_i = f_\theta(x_i)$
                \State Compute loss $L = \frac{1}{B} \sum_{i=1}^{B} \ell(\hat{y}_i, y_i)$
                \State Compute gradients $\nabla_\theta L$
                \State Update parameters $\theta = \theta - \alpha \nabla_\theta L$
               
            \EndFor
        \EndFor
    \State Return trained parameters $\theta$
    \end{algorithmic}
\end{algorithm}
 This section describes the training methodology. The autoencoder training process is detailed in Algorithm \ref{algo:autoencoders}. The training begins by initializing the weights of the autoencoders across different layers. Starting with a pre-trained model, an autoencoder is integrated into the key and value structures at one layer at a time. All the weights except the autoencoder of the layer of interest is frozen during the training process. Key hyperparameters, such as learning rate, batch size B, and number of epochs E, are set. For each mini-batch, predictions are generated, and the loss is calculated based on the cross-entropy loss of the model’s output . Additionally, an L1 norm loss is computed, reflecting the difference between the true and predicted outputs of the autoencoder. This L1 loss, scaled by an empirical value, is added to the original loss. This is highlighted in Algorithm \ref{algo:autoencoders}. Backpropagation is then performed using this combined loss, updating only the autoencoder weights in the specific layer based on the gradients. This process repeats for each mini-batch across all epochs, and the entire procedure is applied iteratively for every layer in the network having an autoencoder.

Once the initial parameters are obtained, they are loaded into the autoencoders in the designated layers for compression. After the weights are applied, the model is fine-tuned over a set number of epochs, freezing all layers parameters except the autoencoder weights. During fine-tuning, the L1 norm is recalculated between the actual and predicted values from the autoencoder in the specific layers, scaled by an empirical value. The sum of all the scaled L1 losses are added to the final loss for back propagation higlighted in Algorithm \ref{algo:autoencoders}. Upon completing the fine-tuning process, the final autoencoder weights can be used in the selected layers for further downstream zero shot tasks which are not finetuned. 

The training methodology for key and value optimization is outlined in Algorithm 2. During each training epoch, the dataset samples are processed, and in each mini-batch iteration, the key and value heads from different layers are recorded. After capturing these values, the L1 norm between consecutive layers is computed. The L1 norm is then averaged across mini-batches. Based on this, an empirical threshold is established to determine which key and value heads should be replaced in their respective layers. After fixing the key and value heads to be replaced from respective layers based on the threshold, a fine tuning process is done on the dataset over a certain number of training epochs. An L1 norm between the actual KV values and the reused KV values for the key and value data structure is computed during each mini batch. The L1 norm loss is scaled and is added to the Cross Entropy loss at the end. This final loss is used in the backpropagation step. The final parameters of the model are saved at the end of finetuning and the model with the updated parameters is used further for zero shot downstream tasks.

\subsection{Quantization} 
\label{sec:Quant}
Quantization can be applied on top of the previously mentioned optimization to achieve additional benefits in terms of memory compression, as shown in Eq. \ref{eq:quant}. One approach is to quantize the full floating-point values to int8 after compression by the encoder. When the key and value data structures are later retrieved from the KV cache, they can be dequantized before passing through the decoder. This process further reduces memory requirements by compressing the values. However, the trade-off is the overhead introduced by the quantization and dequantization processes. This method is most beneficial when memory compression is the primary objective.
\begin{align}
    scale &= 255/(max(x)-min(x)) \nonumber \\
    zeropoint &= -round(scale*min(x))-128 \nonumber \\
    Xquant &= round(scale*x + zeropoint) \nonumber \\
    Xdequant &= (Xquant-zeropoint)/scale \label{eq:quant}
\end{align}

\begin{table*}[!t]
\centering
\caption{\normalsize Results for GPT-2 and TinyLLaMA evaluated on Wikitext, C4, PIQA, and Winogrande across multiple metrics. The table reports baseline performance, compressed performance using autoencoder-based KV compression, and the corresponding KV-cache memory savings.}
\label{tab:autoresults}
\scalebox{1.4}{ 
\begin{tabular}{|c|c|c|c|c|c|}
\hline
\textbf{Model} & \textbf{Benchmark} & \textbf{Metric} & \textbf{Baseline} & \textbf{Compressed} & \textbf{Memory Savings} \\ \hline
\multirow{4}{*}{Tiny LLaMA} & Wikitext & Perplexity & 10.29 & 12.33 (11 layers) & 25\% \\ \cline{2-6} 
                            & C4 & Perplexity & 15.6874 & 16.02 (6 layers) & 13.63\% \\ \cline{2-6} 
                            & Piqa  (zero shot) & Accuracy & 0.6485 & 0.6322 (5 layers) & 11.36\% \\ \cline{2-6} 
                            & Winogrande (zero shot) & Accuracy    & 0.5241
 & 0.513 (22 layers) & 50\% \\ \hline
\multirow{4}{*}{GPT-2}       & Wikitext & Perplexity & 21.4 & 23.3 (10 layers) & 41.6\% \\ \cline{2-6} 
                            & C4 & Perplexity & 34.61 & 37.3 (4 layers) & 25\% \\ \cline{2-6} 
                            & Piqa  (zero shot) & Accuracy & 0.6262 & 0.6055 (10 layers) & 41.6\% \\ \cline{2-6} 
                            & Winogrande (zero shot) & Accuracy    & 0.5083 & 0.5067 (10 layers) & 41.6\% \\ \hline
\end{tabular}
} 
\end{table*}
\section{Results} 
\label{sec:Results}

This section outlines the results of our model evaluations. We utilized two models for functional assessment: GPT-2, a 774-million-parameter model, and TinyLlama, a 1.1-billion-parameter model. The models were evaluated on a range of benchmarks, including WikiText, C4, PIQA, and Winogrande.

WikiText is a language modeling dataset comprising 100 million tokens curated from high-quality, featured Wikipedia articles. C4 (Colossal Clean Crawled Corpus) is a large-scale dataset derived from a cleaned version of the Common Crawl web corpus. It is often used in pretraining large language models. Due to computational limitations, we employed only a small subset of C4 for both training and evaluation in this study.
PIQA (Physical Interaction Question Answering) is a commonsense reasoning benchmark that focuses on the physical world, specifically testing a model's ability to reason about how everyday physical tasks are accomplished. The task consists of multiple-choice questions where the model is required to choose the more plausible answer for performing a given task. Winogrande is a benchmark designed to evaluate commonsense reasoning with a focus on resolving ambiguous pronouns in sentences. It extends the original Winograd Schema Challenge by increasing both the dataset size and the variety of linguistic structures. Each question presents a context followed by a pronoun reference ambiguity and two answer options. The model must infer the correct antecedent based on commonsense understanding.Both PIQA and Winogrande are zero-shot tasks, meaning that the models were evaluated without fine-tuning on these specific datasets. The ability to generalize to these tasks demonstrates the model's capacity for applying pre-existing knowledge to new, unseen scenarios. The fine-tuning for the C4 dataset is based on the saved weights from separately fine-tuning autoencoders on the Wikitext dataset.

\subsection{Accuracy Evaluation}

 Table~\ref{tab:autoresults} presents the application-level results of the two models across multiple benchmarks. It reports perplexity for Wikitext and C4, and accuracy for PiQA and Winogrande, each compared against their respective baselines. Lower perplexity indicates better language modeling ability, while higher accuracy reflects stronger task performance. The reported memory savings correspond to the reduction in KV cache size obtained through compression of the key and value embedding vectors.

As shown in Table~\ref{tab:autoresults}, different models and datasets exhibit varying tolerance to compression, often with minimal loss in performance. For the TinyLlama model, autoencoders can be applied to up to 11 layers, compressing the key and value vectors by a factor of two without significant change in perplexity. In contrast, for the C4 dataset, only up to 6 layers can be compressed before the perplexity begins to increase noticeably. 

In zero-shot evaluation tasks, PiQA maintains stable accuracy with compression applied to 5 layers, while Winogrande tolerates compression across all 22 layers, achieving nearly 50\% reduction in KV cache memory. Similarly, for GPT-2, up to 10 layers can be compressed for Wikitext, PiQA, and Winogrande with negligible accuracy loss, whereas for C4, only 4 layers can be compressed effectively.

Overall, these results demonstrate a clear trade-off between accuracy retention and achievable memory savings. The tolerance to compression depends strongly on the dataset and task characteristics, emphasizing that optimal compression levels should be selected based on application-specific accuracy requirements.

\begin{table}[t]
\centering
\caption{\normalsize Perplexity results for GPT-2 on Wikitext under different levels of key and value head replacement. The table reports baseline perplexity, compressed perplexity, and the corresponding KV-cache memory savings achieved by the replacement strategy.}
\normalsize 
\label{tab:keyvalue}
\renewcommand{\arraystretch}{1.5} 
\begin{tabular}{|c|c|c|c|}
    \hline
    \textbf{Baseline} & \textbf{Compressed} & \textbf{heads replaced} & \textbf{savings} \\ \hline
    21.4  & 30.8  & All key and value  & 50\% \\ \hline
    21.4  & 26.4  & all key  & 25\%  \\ \hline
    21.4 & 26.4 & all value & 25\% \\ \hline
    21.4 & 21.8 & 19 key & 6.59\% \\ \hline
    21.4 & 23.32 & 25 value & 8.68\% \\ \hline
    21.4 & 23.9 & 36 key and value & 12.5\% \\ \hline
    
\end{tabular}
\end{table}

\begin{table}[t]
\centering
\caption{\normalsize GPT-2 results on Wikitext for different amounts of key and value head replacement, both alone and in combination with autoencoder-based KV compression. The table reports baseline and compressed perplexity values along with the resulting KV-cache memory savings, showing that combining head replacement with autoencoder compression yields the highest reduction in memory footprint.}
\label{tab:combined}
\normalsize 
\renewcommand{\arraystretch}{1.5} 
\resizebox{\linewidth}{!}{
\begin{tabular}{|c|c|c|c|}
    \hline
    \textbf{Dataset} & \textbf{Baseline} & \textbf{Compressed} & \textbf{Memory Savings} \\ \hline
    wikitext  & 21.4 (Perpl) & 23.9  & 12.5\% (heads)\\ \hline
    wikitext  & 21.4 (Perpl)  & 23.9  & 47.85\% (aut+heads)\\ \hline
    piqa & 0.6262 (Acc) & 0.5892 & 12.5\% (heads)\\ \hline
    piqa & 0.6262 (Acc) & 0.5936 & 47.85\% (aut+heads)\\ \hline
\end{tabular}
}
\end{table}

The second optimization focuses on the replacement of key and value heads within the attention layers. This technique is first evaluated independently to analyze its individual contribution to model accuracy. Table~\ref{tab:keyvalue} summarizes the results for GPT-2 on the WikiText dataset, illustrating how the model’s accuracy changes as the proportion of replaced heads increases. When nearly half of the layers have their key and value heads replaced, corresponding to around 50\% compression, a noticeable increase in perplexity is observed. This indicates a clear trade-off between compression and predictive quality, as excessive replacement begins to degrade model performance.

The last three columns of Table~\ref{tab:keyvalue} show cases where only a limited number of heads—19 key heads, 25 value heads, and 36 key–value pairs—are replaced. In these selective configurations, replacement is guided by similarity checks performed during the initial analysis. By targeting only the most redundant heads, the model maintains high accuracy while still achieving moderate compression. The perplexity increase remains minimal, demonstrating that selective head replacement can preserve the model’s representational capacity even under compression.

Building on these results, we evaluate two configurations in Table~\ref{tab:combined}:
(1) head replacement only, and
(2) head replacement combined with autoencoder-based compression.
Head replacement alone provides approximately 12.5\% memory savings, while integrating autoencoders increases total savings to 47.85\% for GPT-2 on WikiText. These reductions come with only marginal changes in perplexity or accuracy across both WikiText and PIQA. The results show that combining selective head replacement with strategically placed autoencoders offers substantially greater compression than either approach alone, while preserving model quality relative to the baseline.

\begin{table}[t]
\centering
\caption{\footnotesize Accuracy on PIQA for GPT-2 and TinyLLaMA under baseline, autoencoder (AE), and autoencoder+quantization (AE+Q) compression.}
\label{tab:int8}
\footnotesize
\renewcommand{\arraystretch}{1.15}
\begin{tabular}{|p{2.4cm}|c|c|c|}
\hline
\textbf{Model / Task} & \textbf{Base} & \textbf{AE} & \textbf{AE+Q} \\ \hline

GPT-2 PIQA (10L) & 0.6262 & 0.6055 & 0.6039 \\ \hline
TinyLLaMA PIQA (5L) & 0.6485 & 0.6322 & 0.6219 \\ \hline

\end{tabular}
\end{table}
We further evaluate the impact of incorporating quantization into this hybrid compression pipeline. In this setup, an Int8 quantization was applied using the formulation given in Eq.~\ref{eq:quant}. Table~\ref{tab:int8} reports the baseline performance, the autoencoder-compressed performance, and the combined autoencoder + Int8 quantized results. The results show that Int8 quantization introduces only negligible accuracy degradation for both GPT-2 and TinyLLaMA on PIQA. This demonstrates that quantization can be effectively stacked on top of autoencoder-based KV compression to provide additional memory reduction without compromising model quality.

These results confirm that the hybrid approach can achieve significant memory reduction while maintaining model accuracy. By selectively choosing specific heads and layers for autoencoder placement, the model achieves efficient compression without any degradation in performance. This hybrid compression framework can also be extended to integrate other techniques such as quantization or pruning. Together, these methods offer a flexible and scalable solution for enabling memory-efficient inference across different transformer architectures.

\begin{figure}[t]
  \centering
  \includegraphics[width=\columnwidth]{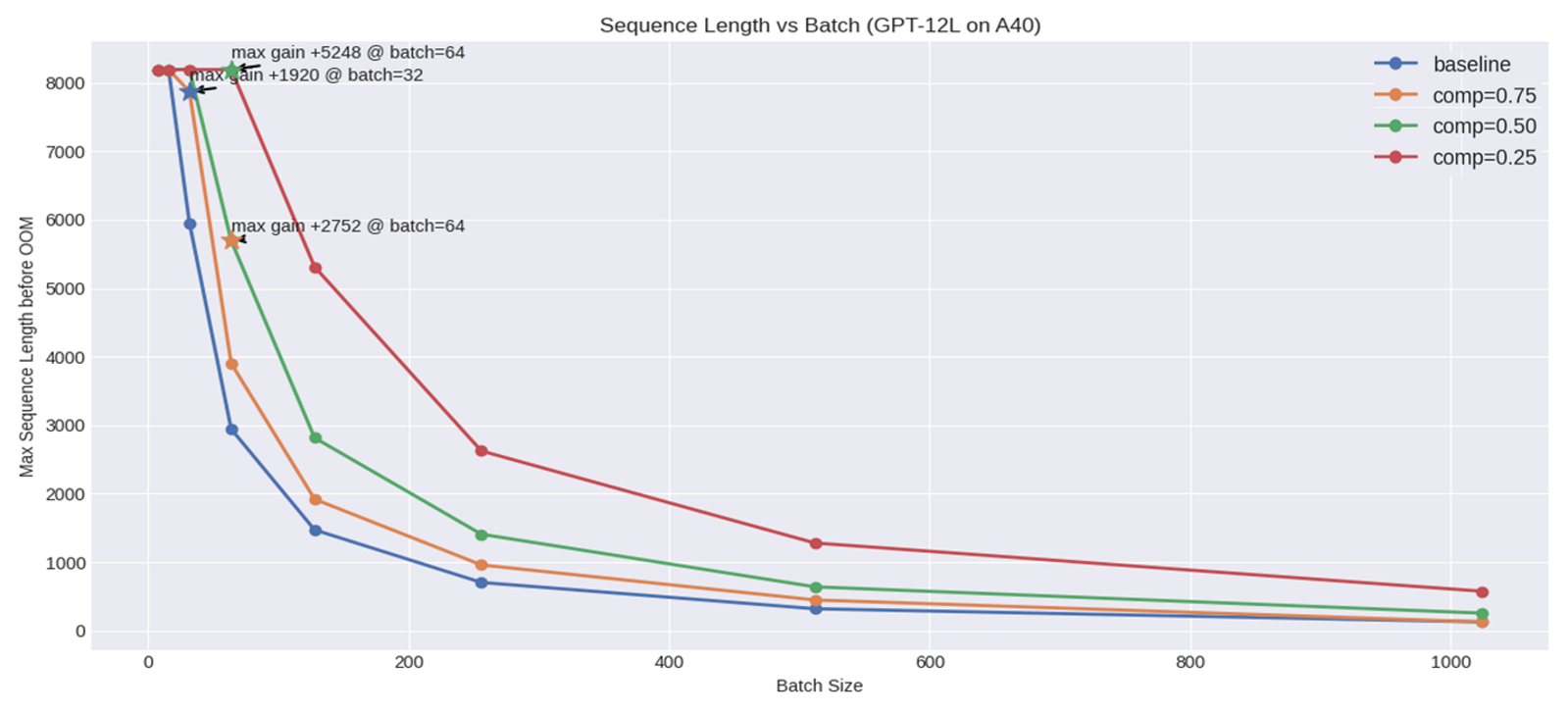}
  \caption{System-level analysis of maximum achievable sequence length as a function of batch size for GPT-2 on an NVIDIA A40 GPU under different KV-cache compression levels. Higher compression rates enable either longer context lengths at fixed batch size or larger batch sizes at fixed sequence length before running out of GPU memory.}
  \vspace{-4mm}
  \label{fig:gptseq}
\end{figure}
\begin{figure}[t]
  \centering
  \includegraphics[width=\columnwidth]{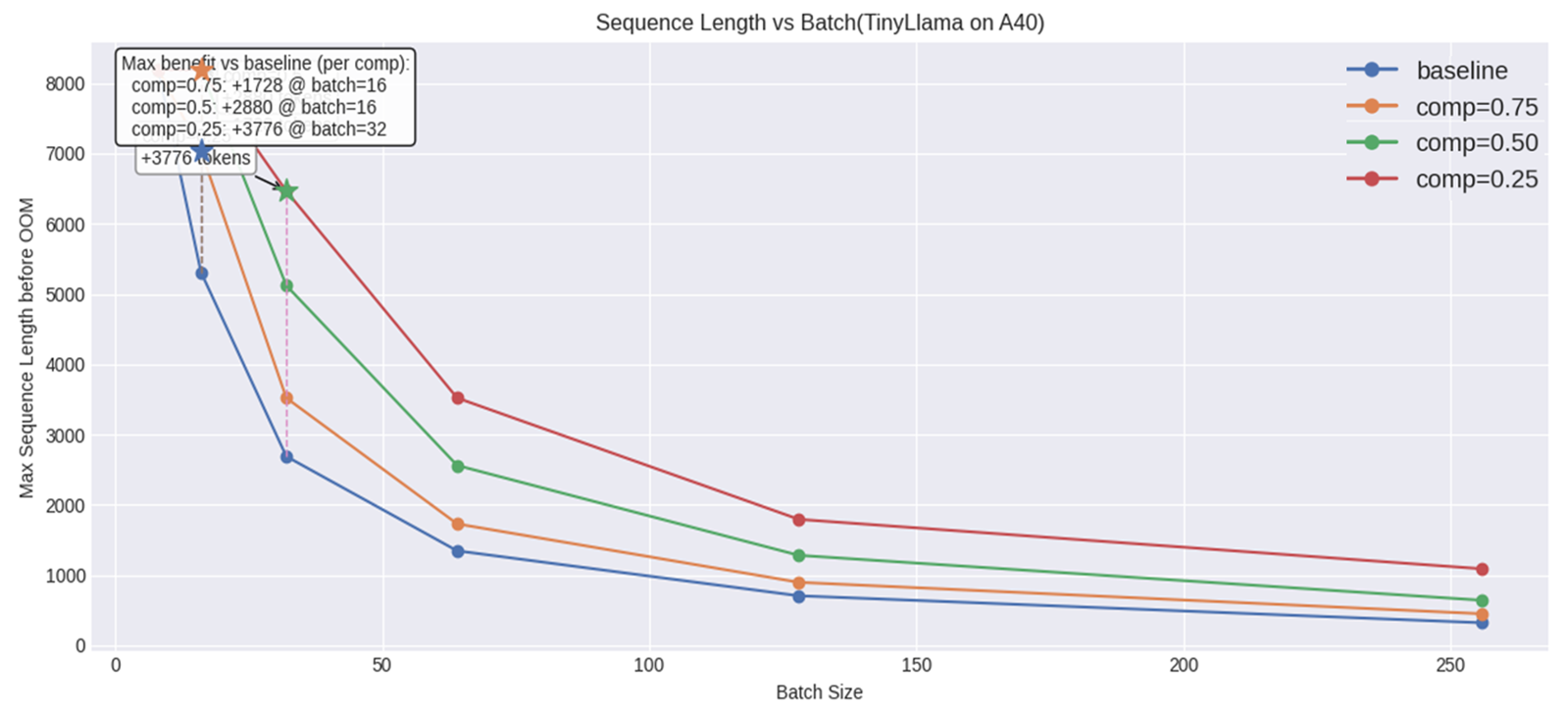}
  \caption{System-level analysis of maximum achievable sequence length as a function of batch size for Tiny Llama on an NVIDIA A40 GPU under different KV-cache compression levels. Higher compression rates enable either longer context lengths at fixed batch size or larger batch sizes at fixed sequence length before running out of GPU memory.}
  \vspace{-4mm}
  \label{fig:tinyllamaseq}
\end{figure}

\subsection{System Evaluation}
A NVIDIA A40 GPU was used for the system-level evaluation of two transformer models, GPT-2 and TinyLlama. Figure~\ref{fig:gptseq} presents the maximum sequence length achievable for different batch sizes when running GPT-2 on the A40 GPU before encountering an out-of-memory (OOM) condition. The four curves in the figure represent different levels of compression applied to the key–value (KV) tensors. The lowest curve corresponds to the baseline configuration without compression, whereas the upper curves correspond to progressively higher compression ratios.

At a fixed batch size, applying a higher compression ratio enables the model to process longer sequences before reaching the GPU memory limit. Similarly, at a fixed sequence length, increasing the compression ratio allows for a larger batch size, leading to better GPU utilization and higher overall throughput. For GPT-2, the results show a clear improvement: the maximum sequence length increases by 5248 tokens for a batch size of 64 with 75\% compression, by 2752 tokens for a batch size of 64 with 50\% compression, and by 1920 tokens for a batch size of 32 with 25\% compression compared to the baseline.

A similar trend is observed for TinyLlama, as shown in Figure~\ref{fig:tinyllamaseq}. With 75\% compression, the model achieves a maximum sequence length gain of 3776 tokens at a batch size of 32. For 50\% compression, the gain is 2880 tokens at a batch size of 16, and for 25\% compression, it is 1728 tokens at a batch size of 16. These results confirm that higher KV-cache compression consistently improves the achievable sequence length and batch size across both models, without compromising accuracy or model quality.

\section{Conclusion}

As LLMs continue to scale in size and context length, the KV cache has become
a dominant source of memory consumption during autoregressive decoding. This
growing footprint limits both sequence length and batch size, creating practical
constraints for efficient inference on modern hardware.

To address this challenge, we proposed two complementary techniques: a learned 
autoencoder that compresses key and value vectors along the embedding dimension, 
and a similarity-based KV head reuse mechanism that removes redundant structures 
across layers. These methods operate without architectural modification and can 
be integrated directly into existing transformer models.

Across GPT-2 and TinyLLaMA, our approach yields up to 47.85\% KV-cache memory 
reduction with minimal degradation in perplexity or zero-shot accuracy. System 
measurements further show that the reduced KV footprint translates into longer 
achievable sequence lengths and larger batch sizes before OOM on GPU hardware.

Overall, embedding-dimension compression combined with selective head reuse 
offers a practical and general solution for memory-efficient LLM inference, and 
can be used alongside quantization, pruning, or architectural variants for 
additional gains.

\bibliographystyle{unsrt}
\bibliography{paper}

@article{dettmers2024gptq,
  title        = {GPTQ: Accurate Post-Training Quantization for Generative Pretrained Transformers},
  author       = {Dettmers, Tim and Lewis, Mike and Shleifer, Sam and Zettlemoyer, Luke},
  journal      = {Transactions on Machine Learning Research},
  year         = {2024},
  url          = {https://openreview.net/forum?id=tY0NdxW7i1}
}

@article{awq2024,
  title        = {AWQ: Activation-aware Weight Quantization for LLM Compression and Acceleration},
  author       = {Lin, Ji and Tang, Yuxin and Yuan, Zechun and Han, Song},
  journal      = {arXiv preprint arXiv:2306.00978},
  year         = {2024},
  url          = {https://arxiv.org/abs/2306.00978}
}

@inproceedings{shazeer2023gqa,
  title        = {Efficient Transformers with Grouped Query Attention},
  author       = {Shazeer, Noam and Ainslie, Joshua and Hsu, Kevin and et al.},
  booktitle    = {International Conference on Machine Learning (ICML)},
  year         = {2023},
  url          = {https://arxiv.org/abs/2305.13245}
}

@article{leviathan2024mistral,
  title        = {Mistral 7B: Open-Weights Language Models at Scale},
  author       = {Leviathan, Yaniv and Lin, J{\'e}r{\'e}my and Bikel, Daniel and others},
  journal      = {arXiv preprint arXiv:2310.06825},
  year         = {2024},
  url          = {https://arxiv.org/abs/2310.06825}
}

@article{li2024kvprune,
  title        = {Dynamic KV Cache Pruning for Efficient Transformer Inference},
  author       = {Li, Zeyu and Zhang, Wei and Wang, Peng and others},
  journal      = {arXiv preprint arXiv:2403.11250},
  year         = {2024},
  url          = {https://arxiv.org/abs/2403.11250}
}

@article{zhao2024paging,
  title        = {Efficient Key-Value Cache Paging and Offloading for Long-Context Transformer Inference},
  author       = {Zhao, Ming and Chen, Yuhong and Zhou, Jie and others},
  journal      = {arXiv preprint arXiv:2402.01030},
  year         = {2024},
  url          = {https://arxiv.org/abs/2402.01030}
}

@inproceedings{dettmers2023gptq,
  title={GPTQ: Accurate Post-Training Quantization for Generative Pretrained Transformers},
  author={Dettmers, Tim and Lewis, Mike and Shleifer, Sam and Zettlemoyer, Luke},
  booktitle={Transactions on Machine Learning Research},
  year={2023}
}

@article{lin2023awq,
  title={AWQ: Activation-Aware Weight Quantization for LLM Compression and Acceleration},
  author={Lin, Ji and Tang, Yuxin and Yuan, Zechun and Han, Song},
  journal={arXiv preprint arXiv:2306.00978},
  year={2023}
}

@article{rotatekv2024,
  title={RotateKV: Maximizing Quantization Robustness for LLM KV Caches},
  author={Xu, Rui and Fan, Xia and Liang, Zhenyu and Wang, Yufan and Chen, Ziyu and Li, Yitan and Shi, Bokai and Xiao, Hailong and Wang, Gelei},
  journal={arXiv preprint arXiv:2408.10417},
  year={2024}
}

@inproceedings{adnan2024keyformer,
  title={Keyformer: KV Cache Reduction Through Key Tokens Selection for Efficient Generative Inference},
  author={Adnan, Muhammad and Arunkumar, Akhil and Jain, Gaurav and Nair, Prashant J. and Soloveychik, Ilya and Kamath, Purushotham},
  booktitle={Proceedings of the 7th MLSys Conference},
  year={2024}
}

@article{zeng2025lethe,
  title={Lethe: Layer- and Time-Adaptive KV Cache Pruning for Reasoning-Intensive LLM Serving},
  author={Zeng, Hui and Zhao, Daming and Yang, Pengfei and Hou, Wenxuan and Zheng, Tianyang and Li, Hui and Ji, Weiye and Zhai, Jidong},
  journal={arXiv preprint arXiv:2511.06029},
  year={2025}
}

@article{mustafar2025,
  title={Mustafar: Promoting Unstructured Sparsity for KV Cache Pruning in LLM Inference},
  author={Anonymous},
  journal={arXiv preprint arXiv:2505.22913},
  year={2025}
}

@inproceedings{shazeer2019mqa,
  title={Fast Transformer Decoding: One Write-Head is All You Need},
  author={Shazeer, Noam},
  booktitle={ICLR},
  year={2019}
}

@inproceedings{ainslie2023gqa,
  title={GQA: Training Generalized Multi-Query Transformers},
  author={Ainslie, Joshua and Ontanon, Santiago and others},
  booktitle={International Conference on Machine Learning (ICML)},
  year={2023}
}

@article{brandon2024crosslayerkv,
  title={Cross-Layer Key-Value Sharing for Memory-Efficient Transformer Inference},
  author={Brandon, Samuel and Rao, Animesh and Patel, Rohan and Kumar, Aakash},
  journal={arXiv preprint arXiv:2405.06789},
  year={2024}
}

@article{cai2024pyramidkv,
  title={PyramidKV: Dynamic KV Cache Compression Based on Pyramidal Information Funneling},
  author={Cai, Zefan and Zhang, Yichi and Gao, Bofei and Liu, Yuliang and Liu, Tianyu and Lu, Keming and Xiong, Wayne and Dong, Yue and Hu, Junjie and Xiao, Wen},
  journal={arXiv preprint arXiv:2406.02069},
  year={2024}
}

@inproceedings{lee2024infinigen,
  title     = {InfiniGen: Efficient Generative Inference of Large Language Models with Dynamic {KV} Cache Management},
  author    = {Lee, Woosuk and Lee, Jaeho and Seo, Juhyung and Sim, Jae},
  booktitle = {18th USENIX Symposium on Operating Systems Design and Implementation (OSDI 24)},
  year      = {2024},
  pages     = {93--110},
  publisher = {USENIX Association}
}

\end{document}